\documentclass[conference]{IEEEtran}
\IEEEoverridecommandlockouts

\usepackage{cite}
\usepackage{amsmath,amssymb,amsfonts}
\usepackage{algorithmic}
\usepackage{graphicx}
\usepackage{textcomp}
\usepackage{xcolor}
\usepackage{booktabs}
\usepackage{array}
\usepackage{url}

\def\BibTeX{{\rm B\kern-.05em{\sc i\kern-.025em b}\kern-.08em
    T\kern-.1667em\lower.7ex\hbox{E}\kern-.125emX}}

\begin{document}

% ===== TITLE =====
\title{Super-Resolution of Solar Magnetograms via Adaptive Stratified Ensemble Learning
with Uncertainty Estimation}
%\thanks{Identify applicable funding agency here. If none, delete this.}}

% ===== AUTHORS =====
% You can have up to 6 authors. Delete or duplicate blocks as needed.

% \author{}

% \author{
% \IEEEauthorblockN{1\textsuperscript{st} Given Name Surname}
% \IEEEauthorblockA{\textit{dept. name of organization} \\
% \textit{name of organization}\\
% email@institution.edu}
% \and
% \IEEEauthorblockA{\textit{dept. name of organization} \\
% \textit{name of organization}\\
% email@institution.edu}
% }

% \author{
% \IEEEauthorblockN{Anonymous Author(s)}
% \IEEEauthorblockA{
% \textit{Anonymous Affiliation}\\
% Anonymous City, Country\\
% Anonymous email
% }
% }

% \author{
% \IEEEauthorblockN{
% Sina Norouzi Kandalan$^{1}$,
% Jason T. L. Wang$^{2}$,
% Qin Li$^{3}$,
% and Haodi Jiang$^{1,*}$\thanks{*Corresponding author: Haodi Jiang (haodi.jiang@shsu.edu).}
% }
% \IEEEauthorblockA{
% $^{1}$Department of Computer Science,
% Sam Houston State University,
% Huntsville, TX 77341, USA
% }
% \IEEEauthorblockA{
% $^{2}$Department of Computer Science,
% New Jersey Institute of Technology,
% Newark, NJ 07102, USA
% }
% \IEEEauthorblockA{
% $^{3}$Department of Physics,
% New Jersey Institute of Technology,
% Newark, NJ 07102, USA
% }
% }

\author{
\IEEEauthorblockN{
Sina Norouzi Kandalan$^{1}$,
Haodi Jiang$^{1,*}$,
Jason T. L. Wang$^{2}$,
and Qin Li$^{3}$\thanks{*Corresponding author: Haodi Jiang (haodi.jiang@shsu.edu).}
}
\IEEEauthorblockA{
$^{1}$Department of Computer Science,
Sam Houston State University,
Huntsville, TX 77341, USA
}
\IEEEauthorblockA{
$^{2}$Department of Computer Science,
New Jersey Institute of Technology,
Newark, NJ 07102, USA
}
\IEEEauthorblockA{
$^{3}$Department of Physics,
New Jersey Institute of Technology,
Newark, NJ 07102, USA
}
}

% \author{
% \IEEEauthorblockN{
% Sina Norouzi Kandalan$^{1}$,
% Haodi Jiang$^{1,*}$,
% Jason T. L. Wang$^{2}$,
% Qin Li$^{3}$
% }
% \IEEEauthorblockA{
% $^{1}$Department of Computer Science,
% Sam Houston State University,
% Huntsville, TX 77341, USA
% }
% \IEEEauthorblockA{
% $^{2}$Department of Computer Science,
% New Jersey Institute of Technology,
% Newark, NJ 07102, USA
% }
% \IEEEauthorblockA{
% $^{3}$Department of Physics,
% New Jersey Institute of Technology,
% Newark, NJ 07102, USA
% }
% \thanks{*Corresponding author: Haodi Jiang (haodi.jiang@shsu.edu).}
% }

\maketitle

% ===== ABSTRACT =====
\begin{abstract}
Single-image super-resolution of  
Sun's photospheric magnetograms
enables consistent analysis across heterogeneous space-based instruments and supports long-term studies of solar magnetic field evolution. 
We address the 
super-resolution task from SOHO/MDI (low-resolution) 
to SDO/HMI (high-resolution) line-of-sight (LOS) magnetograms using a modified RRDBNet architecture initialized by ESRGAN pretrained weights. 
Through systematic per-image diagnostic analysis, we identify image complexity
as the dominant predictor of reconstruction errors.
To exploit this finding, we introduce an 
adaptive
stratified specialist ensemble 
(SSE) 
of three specialist networks
with uncertainty estimation, 
where each specialist network is 
trained by images from three different complexity strata
using a weighted random sampling strategy. 
During inference, a lightweight router based on input image statistics 
assigns each 
test image to the appropriate specialist network. 
Our experimental results demonstrate the good performance of the proposed ensemble and 
its superiority over 
closely related methods.
\end{abstract}

\begin{IEEEkeywords}
Image super-resolution, 
Deep neural networks, Stratified training, Ensemble learning
\end{IEEEkeywords}

% =====================================================================

\section{Introduction}

\begin{figure}
	\centering
        \includegraphics[width=1\linewidth]{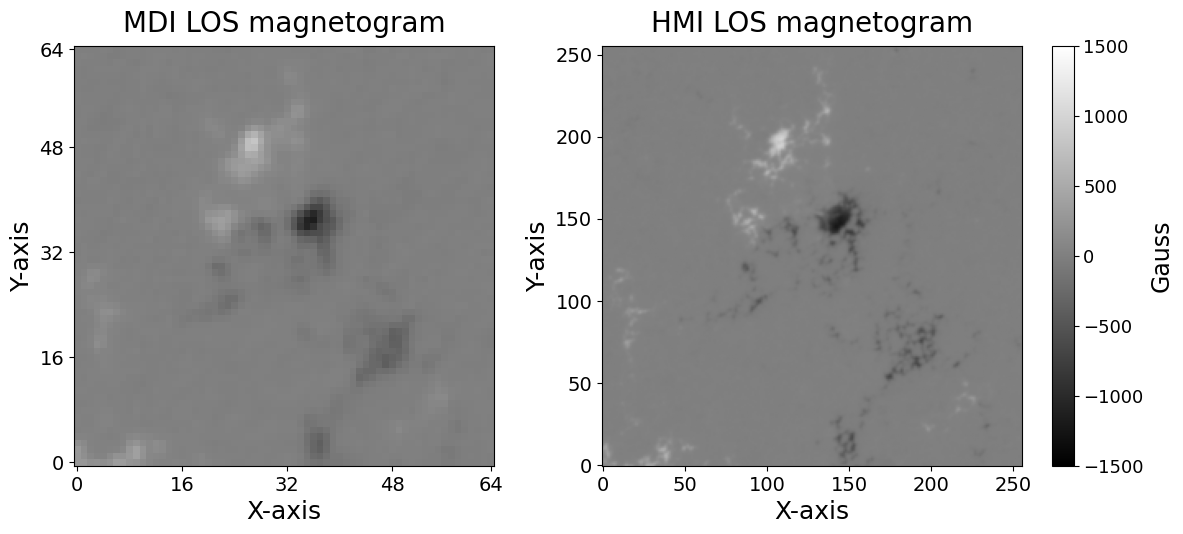}
 	\caption{
    Comparison of MDI and HMI LOS magnetogram images.
    The left panel shows the MDI LOS magnetogram
    observed at 2010-05-01 00:00:00 UT,
    and the right panel shows the corresponding HMI LOS magnetogram 
    observed at the same time.}
	\label{fig:mdi_vs_HMI}
\end{figure}

Sun's photospheric magnetograms---two-dimensional maps of the photospheric line-of-sight (LOS) magnetic field---form the primary observational basis for solar physics studies and space-weather forecasting. The Michelson Doppler Imager (MDI) onboard the Solar and Heliospheric Observatory (SOHO) \cite{scherrer1995} operated from 1996 to 2011 and produced more than fifteen years of continuous magnetogram images at moderate spatial resolution. The Helioseismic and Magnetic Imager (HMI) onboard the Solar Dynamics Observatory (SDO) \cite{pesnell2012,scherrer2012}, which has been operational since 2010, provides magnetogram
images with substantially higher spatial resolution.
Figure \ref{fig:mdi_vs_HMI} compares 
MDI and HMI LOS magnetogram images,
where the resolution of the HMI magnetogram
image 
is four times higher than the resolution of the MDI magnetogram image.
The temporal overlap between MDI and HMI from 2010 to 2011 enables the construction of paired training data for supervised image-to-image translation. A learned $\times 4$ super-resolution mapping from MDI to HMI would extend the HMI-quality dataset backward in time by more than a decade.

In computer vision, 
the enhanced super-resolution generative 
adversarial network (ESRGAN) \cite{wang2018esrgan} and its non-adversarial backbone, 
the residual-in-residual dense block network (RRDBNet), have become standard 
models
for natural-image super-resolution. However, their direct application to scientific imagery encounters two 
domain-specific obstacles. 
First, unlike 
three-channel RGB inputs expected by a pretrained network,
solar magnetograms are single-channel 
signed-magnitude images with bimodal pixel distributions. 
Second, the statistical content of solar magnetograms varies widely across a solar cycle: quiet-Sun regions are nearly featureless, while active regions contain strong magnetic concentrations, sharp field gradients, and frequent flux emergence. 
We hypothesize that this heterogeneity within a dataset is a primary obstacle to reaching the upper performance limit of a single network trained over the full 
pixel
distribution.

To overcome these obstacles, we propose a new approach to super-resolving solar magnetograms.
Our work makes the following contributions. 
\begin{itemize}
\item
We develop a modified RRDBNet architecture with learned input/output adapters and a residual refinement head that enables a transfer from ESRGAN-pretrained RGB weights to single-channel magnetogram super-resolution. 
\item
Through per-image diagnostic analysis, we quantitatively establish image-content complexity as the dominant driver of reconstruction errors. We then design a new framework, named the Stratified Specialist Ensemble (SSE),
that integrates complexity-aware weighted random sampling and 
test-time augmentation to improve solar magnetogram super-resolution.
\item
We introduce a new adaptive metric-gated curriculum loss whose gradient contributions to 
performance metrics
are automatically reweighted as each metric crosses prescribed quality thresholds.
\item
We incorporate uncertainty estimation into the proposed SSE framework
by producing both reconstruction-error-based and 
epistemic uncertainty-based maps. These maps identify structurally complex and error-prone magnetic regions, providing useful confidence estimates for scientific data analysis.
\item 
The proposed SSE framework outperforms closely related methods in three evaluation metrics.
\end{itemize}

The remainder of this paper is organized as follows.
Section \ref{related_work} reviews related work.
Section \ref{data} details the data collection and 
preprocessing procedures.
Section \ref{sec:baseline} describes a baseline method
used in our work.
Section \ref{sec:sse}
describes our SSE framework.
Section \ref{sec:evaluation}
reports the experimental results.
Section \ref{sec:conclusion}
concludes the paper and presents some directions for future research.

% =====================================================================
\section{Related Work}
\label{related_work}

\subsection{Single Image Super-Resolution}
Deep learning super-resolution was pioneered by SRCNN \cite{dong2014}. Subsequent architectures introduced residual connections, recursive structures, and densely connected blocks \cite{huang2017}. 
RRDBNet, the generator component of ESRGAN \cite{wang2018esrgan}, 
integrates dense connections within residual-in-residual blocks, and applies residual scaling to stabilize the training of very deep networks. ESRGAN-class models remain competitive with more recent approaches on natural image 
super-resolution, particularly when computational efficiency is a major concern.

\subsection{Solar Magnetogram Super-Resolution}
Several groups have applied deep learning to super-resolving solar magnetograms.
D\'{i}az Baso and Asensio Ramos \cite{diazbasoasensioramos2018} designed
CNNs with residual blocks to improve SDO/HMI observations.
Rahman et al.~\cite{rahman2020} used a generative adversarial network to
enhance HMI magnetograms, validated against Hinode data.
Xu et al.~\cite{xu2024solarcnn} proposed SolarCNN, an attention-aided CNN
with frequency channel attention (FcaNet) blocks and Mish activations for
the MDI-to-HMI enhancement task.
High-resolution magnetograms, particularly those of flare-producing solar active regions, enable precise tracking of dynamic magnetic structures, capture rapid magnetic-field changes, and provide critical insights into the magnetic processes leading to solar flares.

\subsection{Stratified Training and Adaptive Losses}
Mixture-of-expert methods \cite{jacobs1991adaptive} partition the input space among specialist networks. 
Stratified training has been used in medical imaging \cite{galdran2021balanced}, anomaly detection, and long-tailed recognition \cite{zhang2021distribution}. 
On the other hand,
curriculum learning \cite{bengio2009} trains networks on progressively harder samples; multi-task methods balance several loss terms with adaptive weighting via learned uncertainty \cite{kendall2018}. Our metric-gated curriculum loss adopts a similar philosophy, but couples directly with 
quality metrics, switching the dominant gradient contribution as each metric crosses a target threshold.

% =====================================================================
\section{Data Collection and Preprocessing}
\label{data}

\subsection{Data Acquisition and Pairing}
We used paired magnetogram images from SOHO/MDI 
(low-resolution, LR) and 
SDO/HMI (high-resolution, HR) sampled across the MDI and HMI
overlap period from May 2010 to April 2011. 
The HMI images were taken from the 720-second HMI LOS magnetogram 
series (HMI.M\_720s series)
in Stanford's Joint Science Operations Center (JSOC)
accessible at
\url{http://jsoc.stanford.edu/}.
The MDI observations, 
taken from the 96-minute MDI LOS magnetogram series
(mdi.fd\_M\_96m\_lev182 series)
in JSOC,
were selected by matching their acquisition timestamps
with the timestamps of the corresponding HMI images. 
After pairing the MDI images with HMI images, 
our data set contains 1,569 LR/HR image pairs. 

\subsection{Data Normalization}
Each LOS magnetogram image
was read as a 32-bit floating array. LR patches have $64 \times 64$ pixels while HR patches have $256 \times 256$ pixels, providing a fixed spatial scale factor 
($\times 4$). 
The pixel values were normalized by dividing them by a global scale constant of $1987.34$, chosen to approximate the 99th-percentile absolute pixel value in the training set, and then clamped to $[-1, 1]$.

\subsection{Train-Test Split}
% The 1,569 LR/HR image pairs were then sorted in ascending temporal order. 
% Following the data-splitting strategy used in SolarCNN~\cite{xu2024solarcnn}, the last 76 pairs (most recent, all from April 2011) were reserved as a held-out test set.
% The remaining 1,493 pairs made up the training set. 
% This temporal-tail split is more conservative than a random split because the test images are drawn from a distinct temporal window.
The 1,569 LR/HR image pairs were then sorted in ascending temporal order. 
Following the data-splitting strategy used in SolarCNN~\cite{xu2024solarcnn}, the last 76 pairs (most recent, all from April 2011) were reserved as a held-out test set.
The remaining 1,493 pairs made up the training set. 
A random sample of 10\% of the training data was used for validation.
This temporal-tail split is more conservative than a random split because the test images are drawn from a distinct temporal window.

\subsection{Image-Content Statistics}

\begin{figure}
	\centering
        \includegraphics[width=1\linewidth]{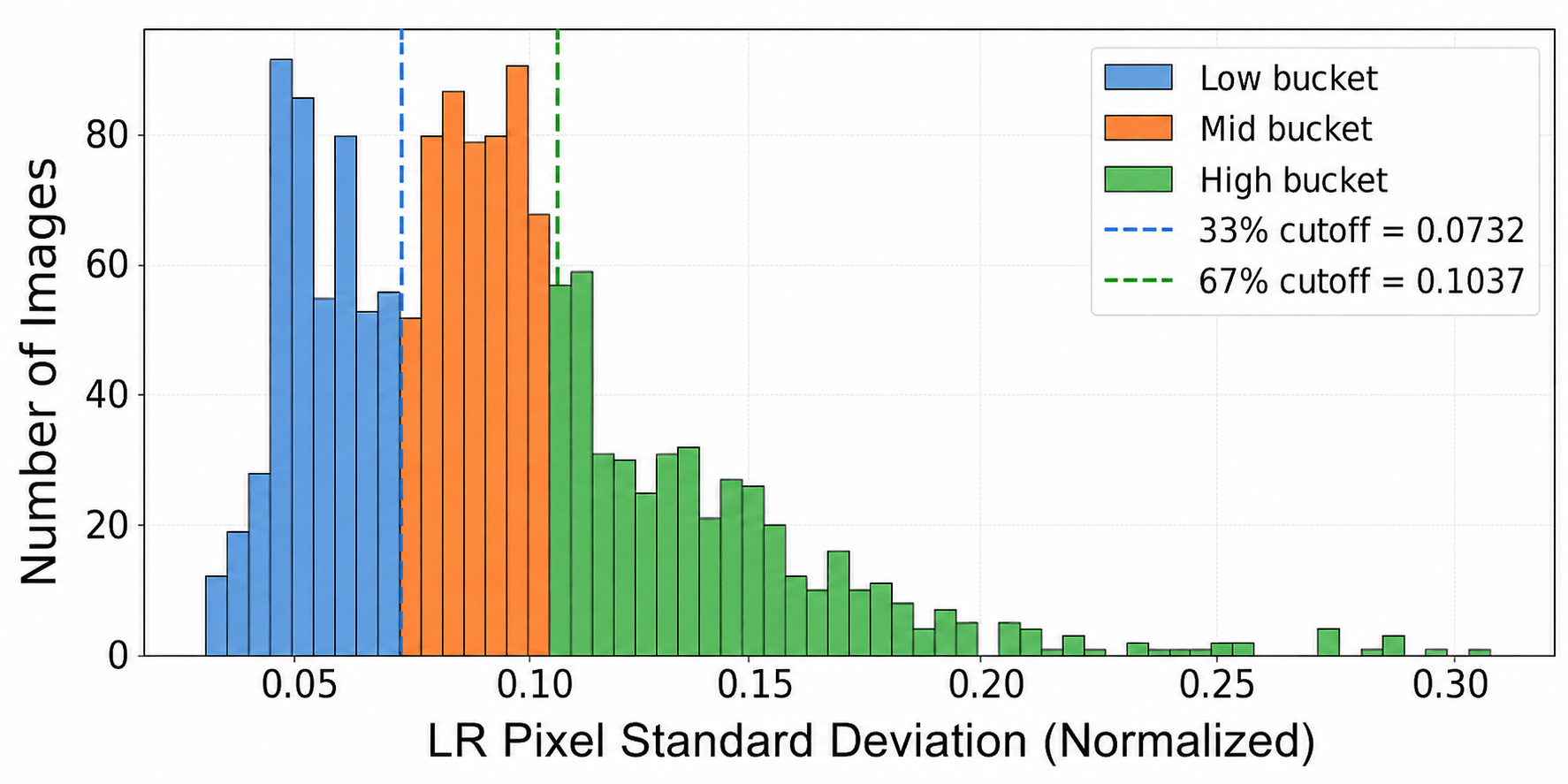}
 	\caption{
Distribution of LR pixel standard deviations over 1{,}493 
    LR training images, partitioned into three complexity buckets at the 33rd
    ($0.0732$, blue dashed line) and 67th ($0.1037$, green dashed line) percentiles.
    The right-skewed tail reflects active-region images with high magnetic
    complexity, motivating stratified training. }
	\label{fig:lr_std_distribution}
\end{figure}

In the training set, we computed the per-image standard deviation of the 
LR
pixel values 
to measure the variability at the image level. 
The standard deviations 
in the 1,493 LR training images
have a median of $0.0889$,
a mean of $0.0951$, 
and range from $0.0310$ to $0.3080$
(see Fig. \ref{fig:lr_std_distribution}).
We divided the 1,493 LR training images 
into three buckets according to their standard deviations: 
low bucket with 493 images ($\mathrm{std}<0.0732$), 
mid bucket with 507 images ($0.0732 \leq \mathrm{std}<0.1037$),
and high bucket with 493 images  ($\mathrm{std}\geq 0.1037$).
% Most LR training images belong to the low and mid buckets.
% Some LR training images in the high bucket 
% have very large standard deviations,
% indicating a skewed distribution. 
% This skewed distribution motivates the 
% design of our SSE framework (see Section~\ref{sec:sse}).
As shown in Fig.~\ref{fig:lr_std_distribution}, the distribution is right-skewed, with a small number of high-variability images extending to much larger standard-deviation values. This variability pattern motivates the design of our SSE framework (Section~\ref{sec:sse}).

\subsection{D4 Augmentation}
\label{D4_Aug}

The MDI--HMI overlap period spans only one year, yielding 1{,}493 training pairs after the temporal-tail split. 
This is a limited data set to 
fine-tune our RRDBNet-based model, which contains approximately $18\times 10^6$
parameters.
To reduce overfitting and improve orientation robustness, we apply the full
dihedral group D4 to every training pair. The D4 group contains eight
transformations, implemented as four rotations combined with two flip states
(no flip and horizontal flip).
The same transformation is applied identically to the LR MDI input and to the HR HMI target. This augmentation is implemented as a virtual $\times 8$ dataset expansion, where each global sample index is decoded into a tuple of pair index and augmentation index. As a result, the effective size of the training set increases from $1{,}493$ to $1{,}493 \times 8 = 11{,}944$ samples without additional data collection.

\section{Baseline Network}
\label{sec:baseline}

\begin{figure}[t]
    \centering
    \includegraphics[width=0.85\linewidth]{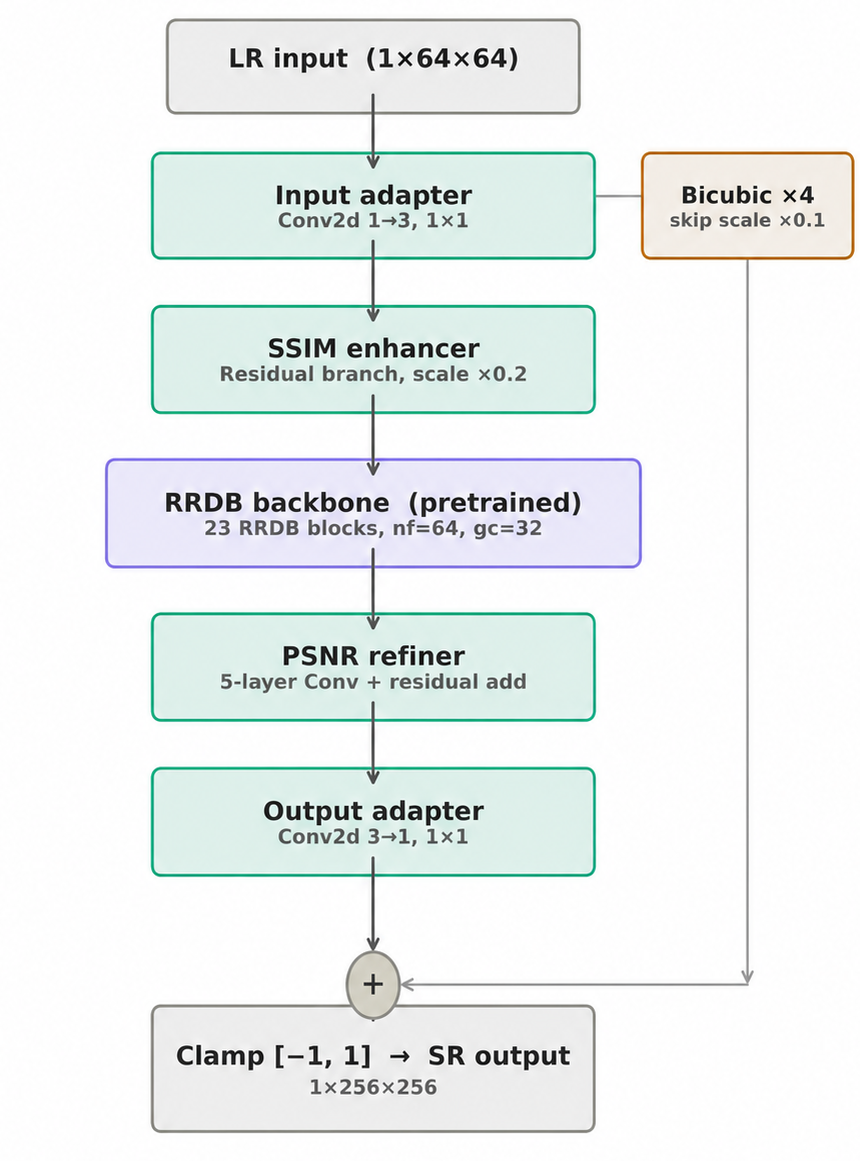}
    \caption{Architecture of the baseline network, which is a modified RRDBNet,
    for $\times 4$ solar magnetogram
    super-resolution. A learned input adapter (Conv2d $1{\to}3$, $1\times 1$)
    maps the single-channel LR magnetogram to 3 channels for the pretrained RRDB
    backbone. 
    An SSIM-enhancer residual block and a PSNR refiner head
    bracket the trunk. 
    A symmetric output adapter projects data back to
    1 channel to produce the super-resolved (SR) output. 
    A bicubic $\times 4$ skip connection scaled by $0.10$ provides
    a conservative bias correction.}
    \label{fig:architecture}
\end{figure}

\subsection{Architecture Overview}
The baseline network is a modified RRDBNet that wraps the 23-block
ESRGAN backbone \cite{wang2018esrgan} and is adapted for single-channel input
through the learned interface modules, as illustrated in Fig.~\ref{fig:architecture}.
A key design decision for the
baseline network
is the deliberate exclusion of the adversarial (GAN)
component of ESRGAN. GAN-based super-resolution encourages the generator to
produce outputs that are perceptually plausible to a discriminator, which in
practice means hallucinating high-frequency texture that is statistically
consistent with natural images but not present in the original observation.
For solar magnetograms, which encode physical measurements of magnetic field
strength, this behavior is scientifically unacceptable: any pixel value that
is not recoverable from the LR input constitutes fabricated physical data.
We therefore use only the RRDBNet generator initialized from the publicly
available ESRGAN $\times 4$ PSNR-oriented weights, inheriting the powerful
pretrained feature representations without the texture-hallucination pathology
of adversarial training. The forward pass is:
\begin{equation}
\hat{y} = \mathrm{clamp}\!\left(g(x) + 0.10 \cdot \mathrm{bicubic}\!\uparrow_4(x),\; -1,\; 1\right),
\end{equation}
where 
$\hat{y}$ is the predicted super-resolved/enhanced MDI magnetogram,
$x$ is the LR input, $g(\cdot)$ is the main network, 
and the second term, namely
$0.10 \cdot \mathrm{bicubic}\!\uparrow_4(x)$,
is a residual bicubic skip connection that acts as a conservative
bias-correction. The operator $\mathrm{clamp}(v, -1, 1) = \max(-1, \min(1, v))$ is 
applied element-wise to every output pixel, constraining the 
prediction to the normalized range $[-1, 1]$.

\subsection{Layer-by-Layer Specification}
All convolutions use a $3\times 3$ kernel with stride 1 and padding 1 unless
otherwise noted. All non-output activations are LeakyReLU with negative
slope $0.2$ except where ReLU is specified.

\textbf{The input adapter ($A_{\mathrm{in}}$)},
consists of
a $1\times 1$ convolution mapping
$1 \to 3$ channels, allowing the pretrained RRDB backbone to be used directly
(6 parameters).

\textbf{SSIM-enhancer block ($S$)}, 
consists of 
a two-layer convolutional block
(Conv $3 \to 64$ + ReLU + Conv $64 \to 3$) applied as a residual update
with scale $0.20$ immediately after the input adapter and before the RRDB
backbone: $h_1 = h_0 + 0.20 \cdot S(h_0)$, with approximately $3{,}500$
parameters. Inspired by the staged decomposition of pixel-fidelity and
structural objectives in \cite{zheng2020ppon} and the dual-module
decoupling philosophy of \cite{sun2025pisasr}, this block operates in the
3-channel feature space at the input resolution, adjusting local contrast
and structural patterns before the pretrained backbone processes them,
encouraging representations more consistent with structural fidelity as
measured by 
SSIM (Structural Similarity Index Measure) \cite{wang2004ssim}. 
The small residual scale ensures that the correction remains
conservative and does not distort the pretrained feature space.

\textbf{RRDB backbone ($R$)},
consists of
the 23-block Residual-in-Residual Dense Block
(RRDB)
trunk from ESRGAN \cite{wang2018esrgan}. 
Each RRDB contains
three residual dense blocks of five $3\times 3$ convolutions with dense skip
connections (growth-channel 32) and residual scale $0.20$. The trunk implements
$\times 4$ spatial upsampling via two 
nearest-neighbor interpolations
interleaved with refinement convolutions. 
The complete trunk is initialized from the publicly available
ESRGAN $\times 4$ weights.

\textbf{PSNR refiner ($P$)},
consists of
a five-layer residual refinement head applied
at the full output resolution after the RRDB backbone: Conv
$3{\to}64{\to}64{\to}64{\to}32 \to 3$ with ReLU activations and a final
$1\times 1$ projection, applied as $h_3 = R(h_1) + P(R(h_1))$
($\sim$$10^5$ parameters). 
Operating on the 
super-resolved feature map, the refiner learns to reduce the systematic
pixel-level errors left by the backbone---in particular the
regression-toward-the-mean smoothing that directly degrades MSE 
(Mean Squared Error)
and
therefore PSNR 
(Peak Signal-to-Noise Ratio)
\cite{zheng2020ppon}. The residual connection ensures that the
backbone output is preserved and only a learned correction is added,
preventing the refiner from destabilizing the structural quality already
achieved by the trunk.

\textbf{The output adapter ($A_{\mathrm{out}}$)},
consists of
a $1\times 1$ convolution
mapping $3 \to 1$ channel (4 parameters).

The baseline network contains approximately 
$18\times 10^{6}$ parameters in total.
The baseline training objective 
is to optimize a static loss, denoted $\mathcal{L}_{\mathrm{static}}$, that
combines the Mean Squared Error (MSE) of pixels with a
hinged penalty in  
SSIM and a variance-matching term 
for penalizing
solutions that systematically under-represent pixel variance.
Let $\hat{y}$ denote the 
predicted super-resolved/enhanced MDI magnetogram, and 
$y$ denote
the corresponding ground-truth HR HMI magnetogram.
Formally,
\begin{align}
\mathcal{L}_{\mathrm{static}} = \;& \mathrm{MSE}(\hat{y}, y) \nonumber\\
& + \lambda_{\mathrm{SSIM}} \max\!\left(0,\; \mathrm{SSIM}_{\mathrm{floor}} - \mathrm{SSIM}(\hat{y}, y)\right) \nonumber\\
& + \lambda_{\mathrm{var}} \left|\sigma(\hat{y}) - \sigma(y)\right|,
\label{static_loss}
\end{align}
with 
$\lambda_{\mathrm{SSIM}} = 0.10$, $\mathrm{SSIM}_{\mathrm{floor}} = 0.90$,
and $\lambda_{\mathrm{var}} = 0.05$. 
The input to the baseline network is
the LR MDI magnetogram.

\subsection{Training Schedule}
% Baseline training was conducted in three phases, 
% with details explained below,
% over 150 epochs.
The model training was conducted over 100 epochs in two phases.

\textbf{Warmup (epochs 1--25): RRDB frozen.}
% The RRDB backbone contains approximately 
% $17.9 \times 10^6$ of the 
% baseline
% network's
% $18.0 \times 10^6$ parameters---more than $99\%$ of total capacity. 
% When the full
% network is unfrozen from the start, 
% the gradient signal from the small
% adapter and refiner modules is overwhelmed by the sheer number of RRDB
% parameters, and convergence to a good solution is substantially slower.
% By freezing the RRDB trunk for the first 25 epochs and training only the
% lightweight input adapter, SSIM-enhancer, PSNR refiner, and output adapter
% ($\sim$$10^5$ parameters), the network rapidly learns a domain-adapted
% interface between the single-channel magnetogram input and the pretrained
% RGB feature space. This warm-up phase reaches a stable operating point in the
% adapter modules at a fraction of the cost of full-network training and
% provides a well-conditioned starting point for backbone fine-tuning.
The RRDB backbone contains approximately 
$17.9 \times 10^6$ of the $18.0 \times 10^6$ total parameters,
accounting for more than $99\%$ of the network capacity. 
When the full network is unfrozen from the start, 
the small adapter and refiner modules must be optimized jointly with the much
larger pretrained backbone, which can slow early-stage convergence.
Therefore, during the first 25 epochs, the RRDB trunk was frozen and only the
lightweight input adapter, SSIM-enhancer, PSNR refiner, and output adapter
($\sim 10^5$ parameters) were trained. This warmup phase allows the model to
learn a domain-adapted interface between the single-channel magnetogram input
and the pretrained RGB feature space, providing a better-conditioned starting
point for subsequent backbone fine-tuning.

\textbf{Main phase (epochs 26--100): RRDB unfrozen.}
% The backbone is released at a conservative learning rate of $2\times 10^{-6}$,
% while the adapter and refiner heads are trained at $3\times 10^{-6}$. The
% lower backbone learning rate prevents catastrophic forgetting of the
% pretrained representations while allowing the trunk to adapt to the
% magnetogram domain.
The RRDB backbone was then unfrozen and trained at a conservative learning rate
of $2\times 10^{-6}$, while the adapter and refiner heads were trained at
$3\times 10^{-6}$. The lower backbone learning rate helps preserve the
pretrained representations while allowing the trunk to adapt to the
magnetogram domain.

Mixed-precision training, gradient clipping with norm $2.0$, and 
exponential moving average (EMA) weight averaging~\cite{polyak1992}
with decay $0.999$ were used throughout training.

% \textbf{Polishing phase (epochs 131--150).}
% The SSIM 
% (Structural Similarity Index Measure)
% penalty weight is reduced from $0.20$ to $0.10$ to allow fine-grained
% PSNR recovery without destabilizing the structural similarity already achieved.
% Mixed-precision training, gradient clipping at norm $2.0$, and 
% exponential moving average (EMA) 
% weight averaging~\cite{polyak1992}
% with decay $0.999$ were used throughout all phases.

% =====================================================================

\section{Stratified Specialist Ensemble}
\label{sec:sse}

\subsection{Weighted Random Sampling}
\label{sec:wrs}

Our SSE framework employs three specialist networks: low, mid, and high,
all of which have the same architecture as the one in Fig. \ref{fig:architecture}. 
Each specialist network is trained with
training images from all three buckets,
namely 
low, mid, and high,
as defined in Fig. \ref{fig:lr_std_distribution},
with a distinct/different specialist-specific sampling weight vector
via the PyTorch \texttt{WeightedRandomSampler}
\cite{paszke2019}.
During network training, the sampling weights are assigned according to the
bucket order: low, mid, high. 
Specifically, the low, mid, and high specialist networks have weight vectors $[1.00, 0.30, 0.05]$, $[0.30, 1.00, 0.30]$, and
$[0.05, 0.30, 1.00]$, respectively.

\subsection{Adaptive Metric-Gated Curriculum Loss}
\label{sec:adapt}

A limitation of the static loss in Equation~(\ref{static_loss}) 
used by the baseline network
is that its relative emphasis on different quality metrics remains fixed throughout training.
However, pixel fidelity, structural similarity, and
magnetic-field correlation may improve and saturate at different rates. 
To address this limitation, we
introduce a metric-gated curriculum loss with adaptive per-metric weights.
Let $M_{\mathrm{PSNR}}$, $M_{\mathrm{SSIM}}$, and $M_{\mathrm{CC}}$ denote the
running validation metrics, and let $\tau_{\mathrm{PSNR}}$,
$\tau_{\mathrm{SSIM}}$, and $\tau_{\mathrm{CC}}$ denote their prescribed
thresholds. For each metric $m$, we define the soft saturation gate:
\begin{equation}
\gamma_m = \sigma_{\beta_m}(\tau_m-M_m),
\end{equation}
where $\sigma_{\beta_m}(z)=1/(1+e^{-\beta_m z})$
is the logistic sigmoid with metric-specific sharpness $\beta_m$.

The adaptive loss,
used by our SSE framework,
is defined as
\begin{align}
\mathcal{L}_{\mathrm{adapt}} = \;&
\alpha_{\mathrm{PSNR}} \gamma_{\mathrm{PSNR}} \mathcal{L}_{\mathrm{PSNR}}
+ \alpha_{\mathrm{SSIM}} \gamma_{\mathrm{SSIM}} \mathcal{L}_{\mathrm{SSIM}} \nonumber\\
&+
\alpha_{\mathrm{CC}} \gamma_{\mathrm{CC}} \mathcal{L}_{\mathrm{CC}}
+ \lambda_{\mathrm{var}} \left|\sigma(\hat{y})-\sigma(y)\right|,
\label{adaptive_loss}
\end{align}
where
$\mathcal{L}_{\mathrm{PSNR}}=\mathrm{MSE}$, 
$\mathcal{L}_{\mathrm{SSIM}}=1-\mathrm{SSIM}$,
$\mathcal{L}_{\mathrm{CC}}=1-\mathrm{CC}$.
The base weights are set to 
$\alpha_{\mathrm{PSNR}}=1.0$, $\alpha_{\mathrm{SSIM}}=0.5$, and
$\alpha_{\mathrm{CC}}=0.5$. The prescribed thresholds are
$\tau_{\mathrm{PSNR}}=37.5$~dB and
$\tau_{\mathrm{SSIM}}=\tau_{\mathrm{CC}}=0.94$, with gate sharpness
$\beta_{\mathrm{PSNR}}=2.0$ and
$\beta_{\mathrm{SSIM}}=\beta_{\mathrm{CC}}=400$. 
The much larger sharpness on SSIM and CC reflects their compressed $[0,1]$ scale relative
to PSNR, so that each gate transitions over a meaningful range of its own
metric.

With the adaptive loss,
the validation metrics are updated at the end of each epoch using an exponential
moving average with decay $0.9$, which is different from the model-weight EMA
with decay $0.999$ 
during training. 
The gates are recomputed at the start of each epoch and
are held fixed during that epoch; therefore, they act as constants during
backpropagation and do not carry gradients. When a metric is below its
threshold, its gate remains open, and its loss term contributes strongly. As the
metric approaches or exceeds its threshold, the gate gradually closes, reducing
that term's effective weight. Thus, the effective weight of each component is
$\alpha_m\gamma_m$, allowing the relative emphasis among metrics to be
adjusted during training.

\subsection{Test-Time Augmentation}
\label{sec:d4}

Test-time augmentation (TTA), in which multiple forward passes are evaluated
on geometrically transformed copies of the input and the output and are averaged
after inverse transformation, is a well-established technique to improve
predictive accuracy without additional training \cite{shanmugam2021better}.
During inference/testing, our SSE framework also adopts 
the D4 transformation group
as described in Section \ref{D4_Aug}
where D4 is applied to training image pairs.
The D4 augmentation used in the SSE testing consists of
four rotations and two flip states, for a total of eight transformations. For each LR MDI test image $x$, we generate eight transformed inputs $T_i(x)$, where $i=1,\ldots,8$. Each transformed input is passed through the trained super-resolution model $g(\cdot)$. 
The predicted super-resolved/enhanced outputs are then mapped back to the canonical orientation using the inverse transformation $T_i^{-1}(\cdot)$ and averaged pixel-wise:
\begin{equation}
\hat{y}_{\mathrm{TTA}} = \frac{1}{8} \sum_{i=1}^{8} T_i^{-1}\!\left(g(T_i(x))\right).
\end{equation}

This procedure reduces the sensitivity to the particular orientation of the input test image and reduces prediction variance by averaging the eight aligned outputs. Since the D4 transformations preserve the image dimensions and are exactly invertible, they can be applied directly to the LR input and the predicted super-resolved/enhanced output. In our test set, the reduction in variance from TTA provides
a consistent and measurable improvement over single-pass inference.

\subsection{Training and Inference Algorithms}
Our SSE framework consists of three 
specialist networks.
Each specialist is trained in two phases. 
During warm-up, the RRDB backbone
is frozen, and head modules are trained at a higher learning rate. 
During fine-tuning, the backbone is unfrozen at a small learning rate, while the head
learning rate is reduced. 
Mixed precision, gradient clipping at the norm $2.0$,
and $\times 8$ TTA at validation are used throughout.

In inference/testing, each LR input $x$ is routed/assigned
to one of the three specialists based
on the standard deviation of the LR pixel values:
\begin{equation}
\mathrm{route}(x) = \begin{cases}
\mathrm{low}  & \text{if } \mbox{std}(x) < c_{\mathrm{low}}, \\
\mathrm{mid}  & \text{if } c_{\mathrm{low}} \le 
\mbox{std}(x) < c_{\mathrm{high}}, \\
\mathrm{high} & \text{otherwise},
\end{cases}
\label{routingequ}
\end{equation} 
where $c_{\mathrm{low}} = 0.0732$ and 
$c_{\mathrm{high}} = 0.1037$ 
are the
33rd and 67th percentiles of the LR pixel standard deviations over the training set, as illustrated in Fig. \ref{fig:lr_std_distribution}. 
The routed prediction uses $\times 8$ D4 TTA averaging. 

In summary, our SSE framework
consists of three specialist networks:
low, mid, and high.
During training,
each specialist network is trained using training images from all three buckets,
namely 
low,  mid, and high,
with a distinct/different specialist-specific sampling weight vector
as described in Section \ref{sec:wrs},
and the training loss used is
the adaptive loss defined in Equation (\ref{adaptive_loss}).
During inference/testing,
an MDI test image is routed/assigned to one of the three trained specialist networks based on 
Equation (\ref{routingequ}).
The assigned trained specialist network 
predicts a 
super-resolved/enhanced MDI image, which has a resolution close to that of the corresponding HMI 
magnetogram image.

\subsection{Uncertainty Estimation}
\label{sec:uncertainty}

Motivated by recent uncertainty-aware super-resolution studies, 
we
generate four 
uncertainty maps, denoted U1--U4,
to identify regions where the super-resolution reconstruction is less reliable. 
In
single image super-resolution,
pixels with high uncertainty often correspond to challenging
texture or edge regions and have been shown to deserve more attention during reconstruction~\cite{ning2021uncertainty}.
The residual magnitude has been used as a practical uncertainty
estimate~\cite{zhang2025uncertainty}. 

Our four uncertainty maps fall into two
categories: \emph{reconstruction-error maps} 
(U1, U2), which
require ground truth and therefore serve as evaluation-time
references, and \emph{epistemic uncertainty maps} 
(U3, U4), which are
computed from the model's own predictions without any reference image
and are therefore available at deployment, 
where no ground truth
exists.

\subsubsection{Reconstruction-Error Maps}

The first map U1 shows the absolute pixel residual.
Let $X_{ij}$ denote the pixel value at the location whose
coordinate is $(i, j)$ in an image $X$.
Let SR represent a super-resolved/enhanced MDI image, and
let HR represent the corresponding high-resolution
HMI image. Then we define
\begin{equation}
\mathrm{U1}_{ij} = \left|\,\mathrm{SR}_{ij} - \mathrm{HR}_{ij}\,\right|,
\label{eq:abs_residual}
\end{equation}
which directly localizes the pixel-level reconstruction error. 

Following the local SSIM formulation
between a super-resolved/enhanced MDI image and
the corresponding HMI image
\cite{wang2004ssim},
denoted SSIMlocal,
we compute the structural similarity index at
each pixel using Gaussian-weighted local moments over 
an $11\times 11$ window and define
\begin{equation}
\mathrm{U2}_{ij} = 1 - \mathrm{SSIMlocal}_{ij}.
\label{eq:ssim_uncertainty}
\end{equation}
U2$_{ij}$ is close to zero where the reconstruction is locally consistent
with the corresponding HMI image and increases in regions of structural
disagreement, highlighting poorly reconstructed magnetic structures
such as those near active-region boundaries.

\subsubsection{Epistemic Uncertainty Maps} 

The third map U3 is constructed based on the 
TTA standard deviation.
The D4 test-time augmentation (TTA)
described in Section~\ref{sec:d4} already produces eight predictions per image,
which are averaged to form the final output. Instead of retaining only
the mean, we additionally compute the pixel-wise standard deviation
across the eight inverse-transformed predictions. 
Because magnetograms
are rotation- and reflection-invariant on the patch scale, a confident
model should produce the same reconstruction in all eight
orientations.
When the predictions disagree, the output depends on
the orientation of the input, indicating low model confidence. 

The fourth map U4 is constructed on 
the basis of specialist disagreement.
Here, we assign the same LR input (i.e., the test MDI image)
to all three specialist networks and compute the 
pixel-wise standard
deviations across their three outputs. 
The pixel-wise standard
deviations form the pixel values of the U4 map. 
When the three specialist networks agree on a pixel, 
the standard deviation at that pixel is zero, and
hence the corresponding pixel value 
on the U4 map is zero.
When the three specialist networks produce very different predictions for a pixel, the corresponding U4 pixel value becomes larger, indicating greater uncertainty due to specialist disagreement.

% =====================================================================
\section{Experiments and Results}
\label{sec:evaluation}

\subsection{Performance Evaluation and Comparison}

We evaluate the performance of the proposed 
SSE framework
and compare it with closely related methods.
Since the related methods do not have an uncertainty estimation component, uncertainty maps are not considered here.
We adopt three evaluation metrics:
PSNR (Peak Signal-to-Noise Ratio) \cite{hore2010psnr},
SSIM (Structural Similarity Index Measure) \cite{wang2004ssim}, and
CC (Pearson Correlation Coefficient) \cite{liu2022}.
PSNR (dB) is defined as $\mathrm{PSNR} = 10 \log_{10}(\mathrm{MAX}^2 / \mathrm{MSE})$
with $\mathrm{MAX} = 1.0$ on the $[-1, 1]$ scale. 
The implementation of SSIM 
uses an $11\times 11$ Gaussian window
($\sigma = 1.5$) and constants $C_1 = (0.01)^2$, $C_2 = (0.03)^2$
\cite{wang2004ssim}.
CC is computed on flattened, mean-centered tensors per image and
averaged over the test set.

Table~\ref{tab:overall} reports the results of the 76-image test set. 
Among all evaluated methods, the proposed SSE framework achieves the best performance, with a PSNR of $37.66$~dB, SSIM of $0.9443$, and CC of $0.9439$. 
% Compared to the baseline network, SSE improves PSNR from $37.12$ to $37.66$~dB, SSIM from $0.9388$ to $0.9443$, and CC from $0.9359$ to $0.9439$. 
Compared to the baseline network, SSE improves PSNR from $37.12$ to $37.66$~dB, SSIM from $0.9399$ to $0.9443$, and CC from $0.9386$ to $0.9439$. 
SSE also outperforms RRDBNet in all three metrics. Compared with SolarCNN~\cite{xu2024solarcnn}, the proposed SSE framework obtains higher PSNR ($37.66$ vs.\ $37.40$~dB), SSIM ($0.9443$ vs.\ $0.9039$), and CC ($0.9439$ vs.\ $0.8842$). These results demonstrate that the proposed SSE framework provides more accurate super-resolution results than closely related methods.

\begin{table}[h]
\caption{Performance Comparison of Studied Methods} 
% 76-image 
%Test set}
\label{tab:overall}
\centering
\renewcommand{\arraystretch}{1.2}
\begin{tabular}{lccc}
\toprule
\textbf{Method} & \textbf{PSNR (dB)} & \textbf{SSIM} & \textbf{CC} \\
\midrule
SolarCNN \cite{xu2024solarcnn} & $37.40$ & $0.9039$ & $0.8842$ \\
RRDBNet \cite{wang2018esrgan}  & $36.56$ & $0.9306$ & $0.9262$ \\
Baseline Network  & $37.12$ & $0.9399$ & $0.9386$ \\ % Trained with L_static now
% Baseline Network + TTA   & $37.23$ & $0.9396$ & $0.9372$ \\
% SSE$^{-}$   & $37.26$ & $0.9395$ & $0.9378$ \\
% SSE$_{s}$ & $37.31$ & $0.9400$ & $0.9384$ \\
$\mathrm{\textbf{SSE}}$      & $\mathbf{37.66}$ & $\mathbf{0.9443}$ & $\mathbf{0.9439}$ \\
\bottomrule
\end{tabular}
\end{table}

% ===== ADDED FOR R2.7 (per-bucket breakdown) — ALL NEW TEXT IN RED =====
% 7) Could the authors also provide a breakdown of per-bucket performance (i.e., SSE results on low/mid/high complexity test images separately)?

To examine whether the improvement caused by SSE is consistent across
different image complexities, we evaluate the baseline network and SSE
separately on the three 
%test-set 
strata defined by the routing rule in
Equation~(\ref{routingequ}). Using the same cutoffs $c_{\mathrm{low}}$ and
$c_{\mathrm{high}}$ applied at inference, the 76 test images are divided into
17 low-, 28 mid-, and 31 high-complexity images. As shown in
Table~\ref{tab:per_bucket}, SSE outperforms the baseline network in all three
strata and across all three metrics.
The PSNR improvements are $+0.48$,
$+0.46$, and $+0.64$~dB,
the SSIM gains are $+0.0025$, $+0.0044$,
and $+0.0055$,
the CC gains are $+0.0060$, $+0.0052$, and
$+0.0061$,
for the low-, mid-, and high-complexity buckets,
respectively.
%while the corresponding SSIM gains are $+0.0025$, $+0.0044$,
%and $+0.0055$. The corresponding CC gains are $+0.0060$, $+0.0052$, and
%$+0.0061$.
These results indicate that the benefit of SSE is consistent across different
complexity levels.

\begin{table}[h]
\caption{Per-Bucket Comparison of the Baseline Network and SSE}
\label{tab:per_bucket}
\centering
\renewcommand{\arraystretch}{1.2}
\setlength{\tabcolsep}{3.5pt}
\begin{tabular}{lcccccc}
\toprule
& \multicolumn{3}{c}{\textbf{Baseline Network}} 
& \multicolumn{3}{c}{\textbf{SSE}} \\
\cmidrule(lr){2-4} \cmidrule(lr){5-7}
\textbf{Bucket} 
& \textbf{PSNR} & \textbf{SSIM} & \textbf{CC}
& \textbf{PSNR} & \textbf{SSIM} & \textbf{CC} \\
\midrule
Low ($n=17$)
& $39.29$ & $0.9587$ & $0.9339$
& $\mathbf{39.77}$ & $\mathbf{0.9612}$ & $\mathbf{0.9399}$ \\

Mid ($n=28$)
& $37.19$ & $0.9387$ & $0.9338$
& $\mathbf{37.65}$ & $\mathbf{0.9431}$ & $\mathbf{0.9390}$ \\

High ($n=31$)
& $35.86$ & $0.9305$ & $0.9445$
& $\mathbf{36.50}$ & $\mathbf{0.9360}$ & $\mathbf{0.9506}$ \\
\bottomrule
\end{tabular}
\end{table}

\subsection{Ablation Tests}

To evaluate the contribution of each component in the proposed SSE framework,
we conduct ablation tests on  
weighted random sampling ($w$),
the adaptive loss ($a$) and
test-time augmentation ($t$). 
Let $\mathrm{SSE}_{w}$ represent the SSE variant using weighted random sampling only,
$\mathrm{SSE}_{wa}$ represent the variant using
weighted random sampling and the adaptive loss only, and
$\mathrm{SSE}_{wt}$ represent the variant using 
weighted random sampling
and test-time augmentation only.
$\mathrm{SSE}_{wat}$ = SSE.
In variants where $a$ is not included, they are trained using the static loss. 
Note that 
variants without weighted
random sampling ($w$) are 
reduced to baseline-like models 
and are not
considered in ablation tests.

Table~\ref{tab:ablation} summarizes the ablation results. Using weighted
random sampling alone, $\mathrm{SSE}_{w}$ achieves a PSNR of $37.26$~dB, SSIM
of $0.9395$, and CC of $0.9378$. Adding test-time augmentation to this setting,
$\mathrm{SSE}_{wt}$ improves the performance to $37.31$~dB PSNR, $0.9400$
SSIM, and $0.9384$ CC. 
Adding the adaptive loss without test-time augmentation,
$\mathrm{SSE}_{wa}$ 
improves the performance to $37.54$~dB PSNR, $0.9430$ SSIM, and $0.9424$ CC.
The complete SSE framework further improves the results to $37.66$~dB
PSNR, $0.9443$ SSIM, and $0.9439$ CC. 
These results show that each of 
the three components, namely
weighted random sampling, 
the adaptive loss, and
test-time augmentation,
has made a significant contribution to our
SSE framework.

\begin{table}[h]
\caption{Ablation Results of the Proposed SSE Framework} 
\label{tab:ablation}
\centering
\renewcommand{\arraystretch}{1.2}
\begin{tabular}{lccc}
\toprule
\textbf{Method} & \textbf{PSNR} & \textbf{SSIM} & \textbf{CC} \\
\midrule
% $\mathrm{SSE}_{t}$   & -- & -- & -- \\ % it is not SSE, but can be filled later
$\mathrm{SSE}_{w}$   & $37.26$ & $0.9395$ & $0.9378$ \\
% $\mathrm{SSE}_{a}$   & -- & -- & -- \\ % it is not SSE, but can be filled later
$\mathrm{SSE}_{wt}$  & $37.31$ & $0.9400$ & $0.9384$ \\
% $\mathrm{SSE}_{ta}$  & -- & -- & -- \\ % it is not SSE, but can be filled later
$\mathrm{SSE}_{wa}$  & $37.54$ & $0.9430$ & $0.9424$ \\ 
$\mathrm{\textbf{SSE}}$       & $\mathbf{37.66}$ & $\mathbf{0.9443}$ & $\mathbf{0.9439}$ \\
\bottomrule
\end{tabular}
\end{table}

\begin{figure*}
%[!h]
    \centering
    \includegraphics[width=0.99\linewidth]{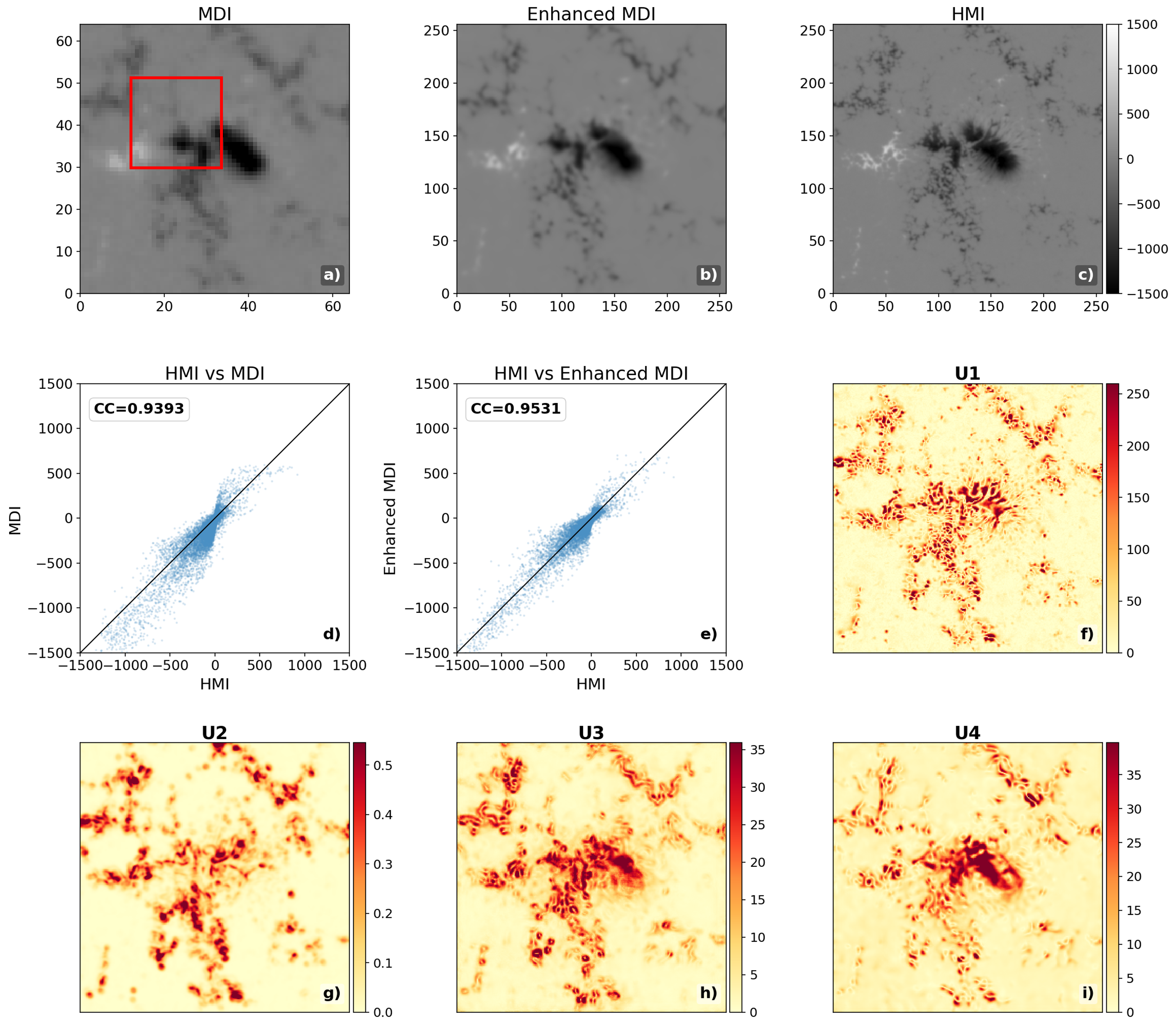}
    \caption{Results of a case study on a
high-complexity test image observed at 2011-04-02 20:48~UT, routed to the high
specialist network. 
% (a)~MDI LR input, 
(a)~MDI LR input, with the red box indicating the region used for the
qualitative 
and quantitative
comparison in Figure~\ref{fig:qual_compare},
(b)~super-resolved/enhanced MDI (SSE
output), (c)~HMI ground truth, (d)--(e)~pixel scatter
plots of HMI vs.\ MDI and HMI vs.\ super-resolved/enhanced MDI,
(f)--(i) four uncertainty
maps.
}
    \label{fig:qualitative}
\end{figure*}

\subsection{A Case Study}

Figure~\ref{fig:qualitative} 
presents the results of a case study, 
showing a high-complexity test image observed
on April 2, 2011 at 20:48~UT, 
which is routed/assigned to the high specialist network.
Panels~(a)--(c) compare the MDI LR input, 
the SSE output (i.e., the 
super-resolved/enhanced MDI image) 
and the HMI ground truth. 
Compared with the LR
input, 
the SSE output recovers sharper magnetic structures and
is visually closer to the HMI reference.
Panels~(d) and (e) show pixel-wise scatter plots against the HMI
reference before and after 
super-resolution.
After super-resolution, the
scatter becomes more concentrated around the 
diagonal 
line, and the
CC increases from $0.9393$ to $0.9531$, 
indicating better
preservation of the field-strength relationship with the HMI reference.
Panels~(f)--(i) show the four uncertainty maps 
described in
Section~\ref{sec:uncertainty}. The reconstruction-error maps, U1 and
U2, highlight regions with 
% fine-scale flux elements and complex magnetic structures, 
abundant small-scale magnetic structures,
where sub-pixel information is difficult to infer from the LR input. 
In contrast, the epistemic uncertainty maps,
U3 and U4, emphasize the 
sunspot 
% core, penumbral structure, 
umbrae and penumbral structures,
which correspond to strong and 
% structurally complex 
highly structured
magnetic features. The spatial correspondence between the
epistemic uncertainty maps and the reconstruction-error maps suggests that
U3 and U4 can provide useful confidence estimates when the HMI
reference is unavailable.

% ===== ADDED FOR R2.4 (qualitative comparison) — ALL NEW TEXT IN RED =====
% NEEDS: Figures/region_20110402_204800_2.png
% 4) No qualitative comparison with prior methods. The case study in Figure 4 shows SSE results but does not include side-by-side visual comparisons with SolarCNN or RRDBNet. Visual comparisons would strengthen the claims of superiority.

\begin{figure*}
\centering
\includegraphics[width=0.88\linewidth]{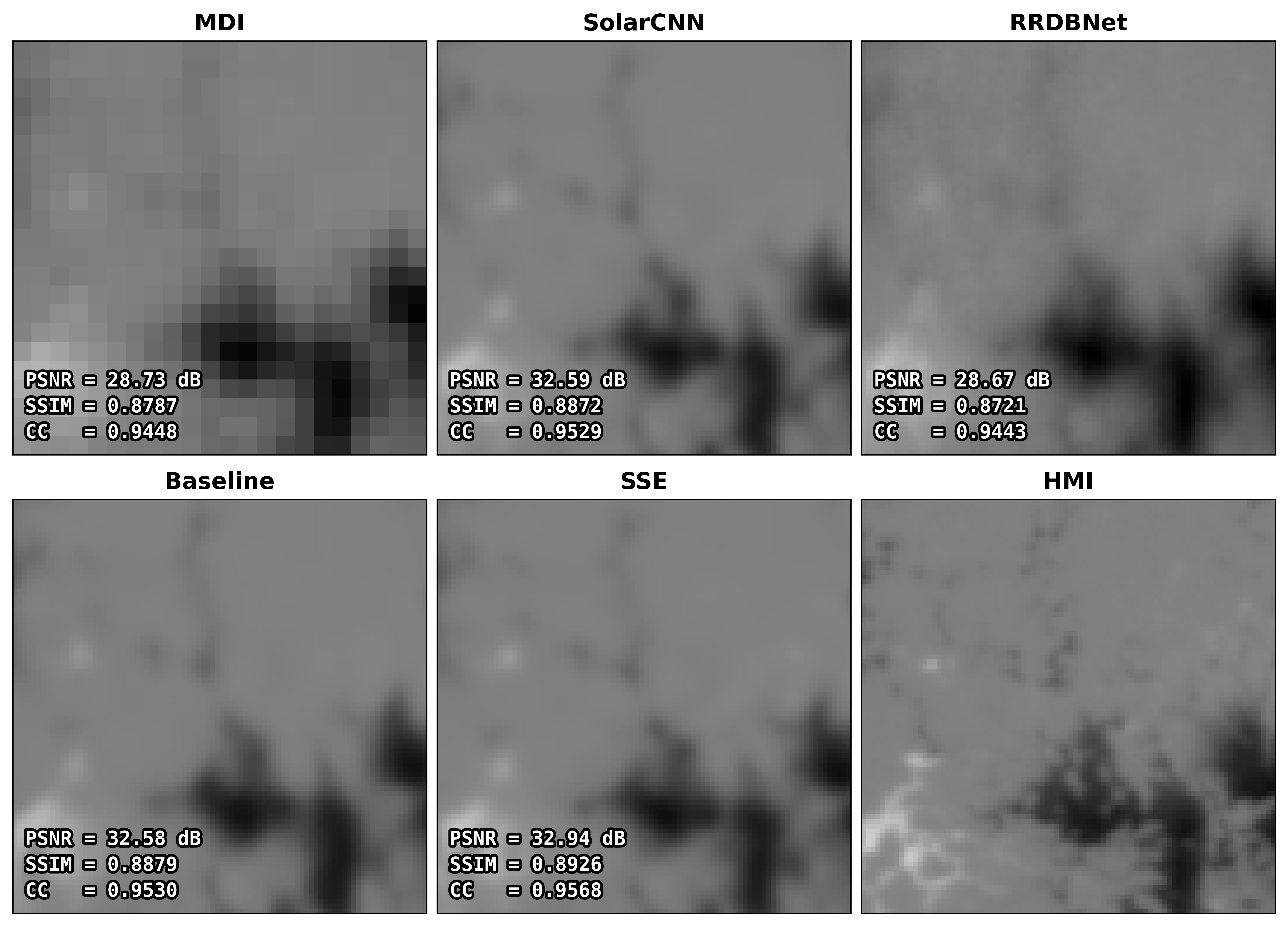}
\caption{Qualitative 
and quantitative
comparison of the zoomed-in region
marked by the red box in Figure~\ref{fig:qualitative}(a). 
Shown in the figure are
 the MDI LR input, SolarCNN
output,
RRDBNet
output, baseline network
output, SSE
output, and HMI ground
truth along with three evaluation metric values. %PSNR, SSIM, and CC with respect to the corresponding HMI region are
%annotated for each super-resolved result.
}
\label{fig:qual_compare}
\end{figure*}

%To further compare SSE qualitatively with related methods,
Figure~\ref{fig:qual_compare} compares the zoomed-in region marked by the red
box in Figure~\ref{fig:qualitative}(a) among the MDI input,
SolarCNN
output,
RRDBNet
output,
baseline
network
output,
SSE
output,
and HMI ground truth. The zoomed-in comparison makes the
differences among the super-resolved results more visible. The MDI input
represents the sunspot group coarsely, while SolarCNN, the baseline
network, and SSE recover sharper magnetic structures and more of the
surrounding fine-scale magnetic features.
Within this zoomed-in region,
SSE achieves the best performance
in all three metrics, with
a PSNR of 
$32.94$~dB, 
an SSIM of
$0.8926$,
and a CC of $0.9568$.
%For comparison, SolarCNN obtains $32.59$~dB, $0.8872$, and
%$0.9529$, while the baseline network obtains $32.58$~dB, $0.8879$, and
%$0.9530$. 
These visual and quantitative results are consistent with the
overall performance comparison in Table~\ref{tab:overall}.

\section{Conclusion}
\label{sec:conclusion}

We presented a Stratified Specialist Ensemble (SSE) framework 
for $\times4$
super-resolution of solar LOS magnetograms. Motivated by the strong
relationship between image complexity and reconstruction difficulty,
%($r=-0.88$ 
%between HR pixel standard deviation and PSNR), 
the SSE routes
an input test image to the low, mid, or high
complexity specialist network based on the LR pixel
standard deviation of the input image. 
Our SSE framework integrates 
three
key components:
stratified training of three specialist networks with
\texttt{WeightedRandomSampler}, 
eightfold D4 augmentation at 
inference,
and an adaptive metric-gated curriculum loss.

These components address different aspects of the magnetogram
super-resolution problem. 
Stratified training allows each specialist to
focus more, by giving a larger sampling weight, on a narrower image-complexity stratum 
via a weighted random sampling strategy.
D4 augmentation improves
robustness to rotations and reflections.
The metric-gated curriculum
loss adaptively balances multiple reconstruction objectives.
Together, these design choices improve the reconstruction fidelity
across different magnetogram-complexity groups.
The per-bucket evaluation further shows that SSE consistently
outperforms the baseline network across the low-, mid-, and high-complexity
strata.

During inference/testing, the proposed SSE framework achieved the best
performance among closely related methods,
including 
SolarCNN \cite{xu2024solarcnn},
RRDBNet \cite{wang2018esrgan} and
the baseline network,
with a PSNR of 
$37.66$~dB, SSIM of $0.9443$, and
CC of $0.9439$. 
%Compared with the baseline network with TTA, the
%adaptive SSE improved PSNR from $37.23$ to $37.65$~dB, SSIM from
%$0.9396$ to $0.9443$, and Pearson CC from $0.9372$ to $0.9439$. 
%The per-bucket results showed consistent improvements across all three
%complexity groups. Compared with SolarCNN~\cite{xu2024solarcnn}, the
%adaptive SSE achieved higher PSNR ($37.65$ vs.\ $37.40$~dB), SSIM
%($0.9443$ vs.\ $0.9039$), and Pearson CC ($0.9439$ vs.\ $0.8842$).
Furthermore, ablation experiments showed that 
all three key components of the SSE framework, namely
weighted random sampling,
the adaptive loss and
test-time augmentation, 
made contributions in improving its overall
performance.

The SSE framework also provides reconstruction-error and 
epistemic
uncertainty maps.
In our case study in which 
we considered a representative high-complexity 
test image, 
the Pearson correlation coefficient (CC) improved from
$0.9393$ for the MDI input to $0.9531$ for the SSE output.
The qualitative 
and
quantitative
comparison with SolarCNN, RRDBNet, and the
baseline network further showed that SSE produced sharper magnetic structures
and achieved the highest PSNR, SSIM, and CC in this case study.
The 
%reference-free 
epistemic
uncertainty maps showed spatial correspondence with
error-prone and 
% structurally complex 
highly structured
magnetic regions, suggesting their
potential usefulness as confidence estimates 
when the HMI reference (ground truth)
is
unavailable. 

Our future work will expand the evaluation beyond the limited
MDI--HMI overlap period, including larger physics-informed synthetic
LR--HR datasets from the
HMI archive \cite{2023SoPh..298...87J, 10903236}, or
perform out-of-sample 
out-of-distribution
validation.
Another direction of future work includes super-resolution of HMI magnetograms using 
even higher-resolution magnetograms such as those from the Big Bear Solar Observatory or Hinode spacecraft.
Further studies concerning whether magnetogram 
super-resolution will
affect the prediction of space weather will also be conducted.

%Haodi: I add ack below.
%Answer: Okay.
\section*{Acknowledgment}
This work was supported in part by 
NASA under grant number
80NSSC25K7708.

% \end{thebibliography}
\bibliographystyle{IEEEtran}
\bibliography{references}

\end{document}